# Graph-Based RL for Sequential Rendezvous Planning with Multiple Patrolling USVs

Seung Woo Han[1,2], Hyeon Bin Yeo[1,3], Seung Heon Oh[1,2*]
[1] Naval Future Innovation Research Group, ROK Navy, Republic of Korea
[2] Seoul National University, Seoul, Republic of Korea
[3] Inha University, Incheon, Republic of Korea
*Corresponding author: suenghuny@snu.ac.kr

*Abstract*— **In patrol missions involving multiple unmanned surface vehicles (USVs), a designated command USV can support fleet management by sequentially rendezvousing with each patrolling USV. These operations require motion-feasible route planning and priority-aware rendezvous sequencing, with the objective of minimizing mission completion time and total tardiness relative to priority-dependent due dates. We propose a framework combining grid-based path planning, surrogate travel-time prediction, and graph-based reinforcement learning. The path planner accounts for ocean currents and approximates turn-rate constraints through multi-stage maneuvers, with path feasibility empirically validated through trajectory-tracking simulations. For rendezvous sequencing, a Feasible Transition Graph (FTG) represents candidate rendezvous events and temporally feasible transitions, enabling a graph neural network to exploit spatiotemporal rendezvous relationships for policy learning. A surrogate predicts travel times from compact current and motion features to avoid repeated path searches during FTG construction and policy training. Evaluation using grid-based path planning shows statistically significant average objective reductions of 10.5%, 3.5%, and 1.5% compared with greedy, MLP-RL, and 2-step lookahead greedy, respectively. The policy achieves these improvements while requiring approximately 1/35 of the latter's computation time. It maintains its advantage over MLP-RL under unseen priority-tier settings, with larger gains as fleet size increases.**

## I. Introduction

Unmanned surface vehicles (USVs) are promising platforms for supporting or replacing crewed vessels in repetitive, resource-intensive missions such as surveillance, environmental monitoring, and search operations. Large-scale, long-duration operations require periodic monitoring and information exchange, but offshore communications may face limitations in coverage, bandwidth, and reliability [1]. One practical approach is to deploy a command USV that periodically rendezvouses with individual USVs for status monitoring and short-range data exchange.

This study considers a maritime surveillance scenario involving a command USV and multiple patrolling USVs. The command USV conducts one rendezvous with each patrolling USV in sequence, while each follows a predefined route. Each patrolling USV has a priority-dependent due date as a soft temporal constraint to encourage earlier visits to higher-priority USVs. The objective is to minimize total tardiness and mission completion time under ocean currents and turn-rate constraints.

This setting creates spatial and temporal dependencies between rendezvous decisions as the targets move along their patrol routes. Ocean currents make travel times dependent on location and direction, while turn-rate constraints expand the path-planning state from $(x, y)$ to $(x, y, \theta)$ by incorporating heading. In particular, the arrival heading affects the travel time and reachability of subsequent rendezvous. These interdependencies complicate rendezvous sequencing and require its integration with detailed path planning.

Previous multi-rendezvous studies used graphs of candidate rendezvous events and feasible transitions to schedule meetings between route-following vehicles and mobile charging vehicles. Solution methods included ILP with a TSP-based heuristic [2] and receding-horizon planning [3]. Related approaches include SMT-based optimization, receding-horizon planning, and dynamic programming [4]-[6]. However, these studies do not consider path planning under marine environmental conditions. In a marine setting, [7] used a current-aware grid to estimate travel energy and temporal transition feasibility, but its position-based representation omitted heading and turning effects. Meanwhile, marine grid-based path planners have considered turn-rate constraints and environmental effects [8], [9], but focus on individual origin-destination pairs without addressing the computational cost of repeated transition evaluations among rendezvous candidates.

Optimization-based rendezvous planning methods [2]-[7] face growing search spaces as the number of targets increases, compounded by repeated path searches to evaluate candidate transitions. Reinforcement learning (RL) has been used to learn policies that rapidly generate visit sequences for static targets, as in the TSP [10], [11]. However, these approaches typically assume static transition costs and feasibility, limiting their application to dynamic rendezvous problems, where transition feasibility and costs vary with position and time.

This study couples current-aware path planning with graph-based RL for priority-aware sequential rendezvous. A grid-based planner approximates continuous turn-rate constraints through multi-stage maneuvers, with allowable turning angles calibrated through trajectory-tracking simulations. To reduce transition evaluation costs, a surrogate-based feasible transition graph (sFTG) represents spatiotemporal relationships using predicted travel times and nominal headings. A GCN encodes the sFTG to learn a rendezvous policy, while surrogate predictions replace repeated path searches during training. The resulting policy is evaluated with grid-based path planning that incorporates ocean currents and turn-rate constraints. To the best of our knowledge, this is the first study to jointly consider these factors in sequential multi-rendezvous planning.

Experiments show statistically significant average objective reductions of 10.5%, 3.5%, and 1.5% compared with greedy, MLP-RL, and 2-step lookahead greedy, respectively.

The proposed method also runs approximately 35 times faster than the computationally intensive lookahead heuristic. Its advantage persists across priority-tier settings and widens over MLP-RL as fleet size increases, supporting the benefits of exploiting graph structure. The main contributions are as follows:

1. A current-aware grid-based path planner with a multi-stage approximation of turn-rate constraints, empirically calibrated through trajectory-tracking simulations.
2. A graph-based RL framework that exploits future rendezvous connectivity, using surrogate predictions to avoid repeated path searches during FTG construction and training.
3. Empirical evidence that the benefit of graph encoding increases with patrol fleet size and persists under grid-based path-planning evaluation.

## II. Problem definition

This section formulates the problem in two stages: current-aware path planning with turn constraints and sequential rendezvous planning.

### A. Current-Aware Path Planning with Turn Constraints

The minimum-time path-planning problem for a USV under ocean currents and turn-rate constraints is formulated in (1)–(8). Let the initial position be $\boldsymbol{X}_s = (x_s, y_s)$, the goal position be $\boldsymbol{X}_f = (x_f, y_f)$, the initial heading be $\theta_s$, the USV speed be $v$, the current field be $\boldsymbol{c}(x, y)$, and the maximum heading rate be $\omega_{\max}$. Headings and bearings are measured counterclockwise from the positive $x$-axis.

$$T_Q(X_s, \theta_s, X_f) := \min_{\zeta(\cdot), T} T \quad (1)$$

$$\text{s.t.} \quad \dot{x}(t) = v \cos\theta(t) + c_x(x(t), y(t)), \;\; t \in [0, T] \quad (2)$$

$$\dot{y}(t) = v \sin\theta(t) + c_y(x(t), y(t)), \;\; t \in [0, T] \quad (3)$$

$$\dot{\theta}(t) = w(t), \;\; t \in [0, T] \quad (4)$$

$$\zeta(t) = \big(x(t), y(t), \theta(t)\big), t \in [0, T] \quad (5)$$

$$|w(t)| \le \omega_{max}, \;\; t \in [0, T] \quad (6)$$

$$(x(0), y(0), \theta(0)) = (x_s, y_s, \theta_s) \quad (7)$$

$$(x(T), y(T)) = (x_f, y_f) \quad (8)$$

Here, $\zeta(t)$ denotes the vessel state trajectory, and $T_Q\big(\boldsymbol{X}_s, \theta_s, \boldsymbol{X}_f\big)$ denotes the minimum travel time. Equations (2)–(8) define the current-affected vessel dynamics, turning constraint, and boundary conditions.

### B. Sequential Rendezvous Planning

The rendezvous planning problem determines the visit sequence of a command USV to $N$ patrolling USVs with hierarchical priorities. The patrol route of USV $i$ is defined as $P_i = \{p_{i,j} \in \mathbb{R}^2 \mid j = 1, \dots, m\}$, where $t_{i,j}^n$ denotes the time at which USV $i$ reaches waypoint $p_{i,j}$ for the $n$-th time, and $d_i$ denotes its due date. Let $\tau$ denote the decision step and $T(\tau)$ the completion time of the $\tau$-th rendezvous. The travel time of the command USV is computed using the minimum-travel-time function $T_Q$ defined above. Let $\pi = (\pi_1, \dots, \pi_N) \in S_N$ denote a permutation representing the visit order of all patrolling USVs. $Q$ returns the arrival-heading associated with the path defined above. The rendezvous planning problem is then formulated in (9)-(15).

$$\min_{\Pi \in S_N} \left[ T(N) + \textstyle\sum_{\tau=1}^{N} max(0, T(\tau) - d_{\pi_\tau}) \right] \quad (9)$$

$$\text{s.t.} \quad T(0) = 0, \; X_c(0) = X_0, \; \theta_c(0) = \theta_0 \quad (10)$$

for each $\tau = 1, .., N$, where $i(\tau) = \pi_\tau$,

$$a_j^\tau = T(\tau - 1) + T_Q(X_C(\tau - 1), \theta_c(\tau - 1), p_{i(\tau),j}) \quad (11)$$

$$n_j^\tau = min\{n : \; t_{i(\tau),j}^{(n)} \ge a_j^\tau\} \quad (12)$$

$$j^*(\tau) = arg \min_j (t_{i(\tau),j}^{(n_j^\tau)} - T(\tau - 1)) \quad (13)$$

$$T(\tau) = t_{i(\tau),j^*(\tau)}^{(n_{j^*(\tau)}^\tau)}, \;\; X_c(\tau) = p_{i(\tau),j^*(\tau)} \quad (14)$$

$$\theta_c(\tau) = Q(X_c(\tau - 1), \theta_c(\tau - 1), p_{i(\tau),j^*(\tau)}) \quad (15)$$

Equation (9) minimizes the final rendezvous completion time and total tardiness across all USVs. In (11), $a_j^\tau$ denotes the earliest arrival time of the command USV at $p_{i(\tau),j}$ at decision step $\tau$. The feasible rendezvous time at $p_{i(\tau),j}$ is determined as the earliest time at which both USVs can be present at the location, allowing the command USV to wait if it arrives before the patrolling USV. Equation (13) selects the earliest feasible rendezvous point $j^*$ along the route of the selected USV. Equations (14) and (15) then update the rendezvous time, position, and command-USV heading. The decision variable is the visit sequence $\pi$, while each rendezvous point is selected to achieve the earliest feasible meeting with the chosen USV.

## III. Modeling

### A. Current-Aware Path Planning with Turn Constraints

This section presents the proposed grid-based marine path-planning framework, emphasizing computational tractability for repeated searches in multi-rendezvous planning (Fig. 1).

#### 1) Grid Graph Construction

The operating area is discretized into a grid, and the USV is assumed to move along straight-line segments between neighboring grid points. As shown in Fig. 1(a), the neighborhood is constructed by adding candidate edges only when they do not intersect existing edges except at the current node. To incorporate turn-rate constraints, each grid point is represented by nodes $v = (X, \theta)$, where $X$ is the position and $\theta$ the arrival heading. The arrival heading depends on the incoming edge, so each grid point has one node per neighboring direction.

For a transition from $\boldsymbol{X}_s$ to $\boldsymbol{X}_f$, the required arrival heading $\theta_{X_s X_f}$ is calculated using (16), where $\bar{c}_\perp$ is the component of the average current between $\boldsymbol{X}_s$ and $\boldsymbol{X}_f$ perpendicular to the transition direction, positive leftward. If $\mid \bar{c}_\perp \mid \ge v$, the transition is excluded because the required heading is infeasible.

$$\theta_{X_s, X_f} = \angle(X_f - X_s) + arcsin(-\frac{\bar{c}_\perp}{v}) \quad (16)$$

A directed edge is created only if the angular difference between the current heading $\theta_s$ and the required heading $\theta_{X_s, X_f}$ does not exceed the maximum allowable angle $\theta_{\max}$.

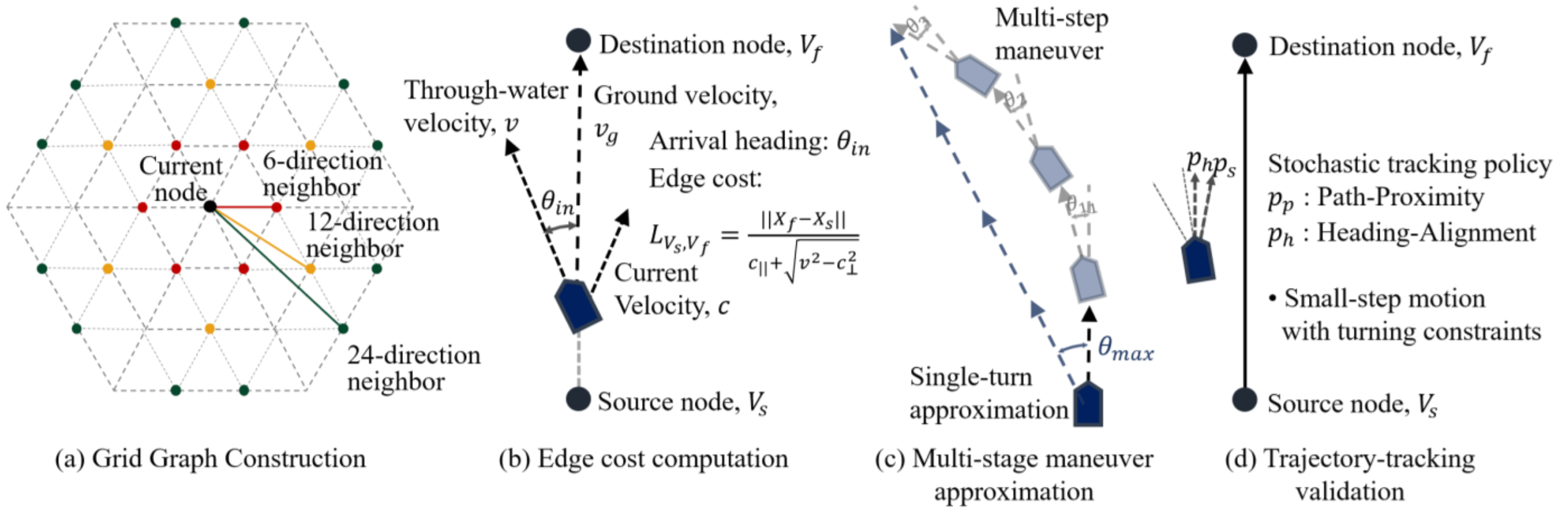

Fig. 1. Grid-based path planning framework under ocean currents and turn-rate constraints

Let $N(\boldsymbol{X}_s)$ denote the neighborhood of $\boldsymbol{X}_s$, and $d_{ang}$ the minimum absolute angular difference between two headings. The feasible neighborhood under this constraint is defined in (17). The edge cost represents the travel time between neighboring grid points under ocean currents, as illustrated in Fig. 1(b), and is calculated using (18), where $\bar{c}_{\parallel}$ is the average current component along the transition direction.

$$N_{\theta_{max}}(X_s, \theta_s) = \{X_f \in N(X_s) \mid d_{ang}(\theta_s, \theta_{X_s, X_f}) \leq \theta_{max}\} \quad (17)$$

$$L_{X_s X_f} = \frac{||\, X_f - X_s \,||}{\bar{c}_{||} + \sqrt{v^2 - \bar{c}_{\perp}^2}} \quad (18)$$

This graph formulation captures both current-dependent travel time and heading-dependent transition feasibility. The grid path is then obtained using standard graph search algorithms. The estimation of $\theta_{\max}$ based on the continuous turn-rate constraint is detailed in the following subsection.

*2) Multi-stage maneuver approximation*

This section introduces a multi-stage maneuver approximation that maps the accumulated displacement from successive small turns to an equivalent straight-line motion along a representative heading, as shown in Fig. 1(c). This approximates continuous turning motion at the grid-transition level. Assuming that heading changes at each small-time step $\Delta t$, the turn-rate constraint is expressed in (19).

$$d_{ang}(\theta_{k+1}, \theta_k) \leq \omega_{max} \Delta t \quad (19)$$

For a maneuver consisting of $N$ time intervals, let $\bar{\theta}$ denote the mean heading and define the heading deviation at step $k$ as $\delta_k = \theta_k - \bar{\theta}$. Assuming that the headings remain close to $\bar{\theta}$ and that $\delta_k$ is sufficiently small, a second-order Taylor expansion yields (20).

$$\sum_{k=1}^{N} \begin{bmatrix} cos\ \theta_k \\ sin\ \theta_k \end{bmatrix} \approx N\left(1 - \tfrac{1}{2} Var(\delta)\right) \begin{bmatrix} cos \bar{\theta} \\ sin \bar{\theta} \end{bmatrix} \quad (20)$$

Equation (20) shows that the accumulated displacement of the multi-stage maneuver is approximately aligned with the mean heading $\bar{\theta}$, while the displacement along that direction decreases as the heading variation increases. Thus, $\bar{\theta}$ can be interpreted as the representative direction of the accumulated motion. This result provides a mathematical basis for the above approximation.

In the absence of ocean currents, the heading required for a given grid transition can be determined geometrically. Under ocean currents, however, determining the allowable heading change becomes more difficult because the accumulated displacement includes current-induced motion at each maneuver stage, as described in (21). Accordingly, $\theta_{\max}$ is empirically calibrated to account for the reachability of multi-stage maneuvers under ocean-current conditions.

$$\Delta X^{(n)} = v \Delta t \sum_{k=1}^{N} \begin{bmatrix} cos\ \theta_k \\ sin\ \theta_k \end{bmatrix} + \Delta t \sum_{k=1}^{N} c(x_k, y_k) \quad (21)$$

*3) Calibration of the Maximum Feasible Turning Angle*

A grid search is conducted over candidate turning angles. For each candidate, trajectory-tracking simulations are performed under varying initial positions, headings, and ocean-current conditions. Path tracking employs a reference strategy that probabilistically selects between a Path-Proximity Strategy, which guides the USV toward the closest point on the reference path, and a Heading-Alignment Strategy, which aligns the USV with the direction of the nearest path segment (Fig. 1(d)). Repeated simulations are used to estimate the tracking success rate $p_{\text{succ}}(\theta)$ for each candidate angle, from which its two-sided 95% Wilson confidence lower bound $p_W^{0.95}(\theta)$ is computed. Using a conservative acceptance threshold of $\eta = 0.99$, the largest candidate angle whose confidence lower bound meets the threshold is selected as the maximum empirically feasible turning angle in (22).

$$\theta_{max} = max\{\theta \in \Theta \mid p_W^{0.95}(\theta) \geq \eta\} \quad (22)$$

This procedure is intended as an empirical calibration of $\theta_{\max}$, rather than a new trajectory-tracking control method.

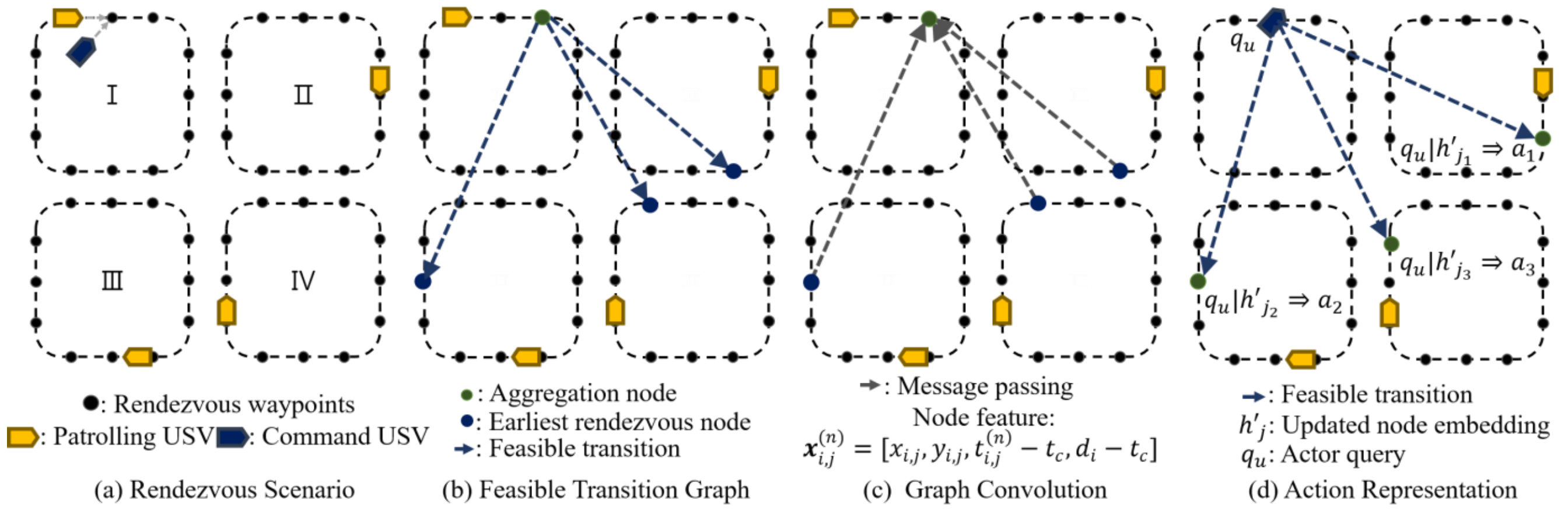

Fig. 2. Graph-Based RL framework for sequential rendezvous planning

The focus is on estimating a grid-level turning constraint from simulation-based tracking performance under continuous turn rate constraints and ocean currents.

#### 4) *Surrogate-based travel-time estimation*

Although the grid-based path planner described above enables efficient graph search, repeatedly invoking it for each candidate rendezvous transition during RL training remains computationally expensive. To alleviate this computational burden, a low-cost surrogate model is introduced to predict USV travel time from low-dimensional features describing the origin–destination geometry and surrounding currents. Current statistics are extracted from inner and outer corridors defined with respect to the straight line connecting the origin and destination. The reference line is used only for feature extraction and does not represent the actual travel path. The resulting feature vector, consisting of current statistics, origin–destination distance, and initial heading information, is defined in (23).

$$z_{od} = \left[\mu_{\perp}^{in}, \left(\sigma_{\perp}^{in}\right)^2, \mu_{\parallel}^{in}, \left(\sigma_{\parallel}^{in}\right)^2, \Delta\theta_{od}, d_{od}, \Delta\mu_{\parallel}^{out-in}, \Delta\mu_{\perp}^{out-in}\right] \quad (23)$$

Here, $\mu_{\perp}^{in}$ and $\mu_{\parallel}^{in}$ denote the mean perpendicular and parallel current components in the inner corridor, with corresponding variances $\left(\sigma_{\perp}^{in}\right)^2$ and $\left(\sigma_{\parallel}^{in}\right)^2$. $\Delta\theta_{od}$ and $d_{od}$ denote the initial heading–bearing difference and origin–destination distance, respectively. The features $\Delta\mu_{\perp}^{out-in}$ and $\Delta\mu_{\parallel}^{out-in}$ represent the differences in the mean current components between the outer and inner corridors, capturing spatial variations in the surrounding current field. The corridor widths are treated as hyperparameters. Furthermore, two prediction targets are considered: the absolute travel time and its residual with respect to the Euclidean reference time $d_{od}/v$.

### B. *Graph-based RL for Rendezvous Planning*

This section presents the RL framework for sequential rendezvous planning. Candidate rendezvous events across multiple patrol routes and their feasible transitions are represented by a Feasible Transition Graph (FTG), which encodes the spatiotemporal connectivity among potential future rendezvous events. The FTG serves as the state representation, and its graph structure is encoded by a graph convolutional network (GCN) [12] to learn an RL policy for rendezvous sequencing. The overall framework is shown in Fig. 2, with its components described using a Markov Decision Process.

#### 1) *State representation*

A node in the FTG represents a waypoint on a patrol route together with the corresponding arrival time. Following the notation in Section II-B, a node is denoted by $v_{i,j}^{(n)} = \left(p_{i,j}, t_{i,j}^{(n)}\right)$. Its feature vector consists of the waypoint position, the remaining time until the patrolling USV reaches the waypoint, and the remaining time until its due date, as defined in (24). Here, $t_c$ denotes the current simulation time.

$$\boldsymbol{x}_{i,j}^{(n)} = [x_{i,j}, y_{i,j}, t_{i,j}^{(n)} - t_c, d_i - t_c] \quad (24)$$

Repeated patrol cycles can generate infinitely many nodes at the same locations, so a computational horizon of the next three arrivals is used to bound the graph size. The FTG includes only unvisited patrolling USVs and is updated over time in both node features and graph structure, forming a time-varying graph. Each FTG node is then connected to the earliest reachable rendezvous candidate of every other patrolling USV. These connections represent the feasible next rendezvous options, as illustrated in Fig. 2(b). The successor node and edge set are defined in (25) and (26), respectively. For each FTG transition, the bearing from departure to destination, $\theta'$, is used as the nominal departure heading. For brevity, $T_Q\left(p_{i,j}, p_{k,l}\right)$ denotes $T_Q\left(p_{i,j}, \theta', p_{k,l}\right)$.

$$v_k^*(v_{i,j}^{(n)}) = \underset{v_{k,l}^{(m)} \in V_k}{argmin}\{t_{k,l}^{(m)} | T_Q(p_{i,j}, p_{k,l}) \le t_{k,l}^{(m)} - t_{i,j}^{(n)}\} \quad (25)$$

$$\boldsymbol{\varepsilon} = \{(v_{i,j}^{(n)}, v_k^*(v_{i,j}^{(n)})) \mid k \ne i\} \quad (26)$$

Conceptually, the FTG is constructed using the travel time and arrival heading obtained from detailed path planning. In practice, however, the estimated travel time from the surrogate model and a nominal heading approximation are used for efficient state generation. Accordingly, the state representation in this study employs a surrogate-based FTG, hereafter denoted as sFTG.

### 2) Graph Convolution

The sFTG is encoded by a GCN that incorporates information from reachable future rendezvous candidates. As illustrated in Fig. 2(c), directed edges represent temporal precedence, allowing each node to aggregate information from its feasible successor nodes. For a node $v = v_{i,j}^{(n)}$, $\mathcal{N}(v)$ denotes the set of reachable successor nodes. For $\mathcal{N}(v) = \emptyset$, $m_v$ is set to 0. The GCN layer is defined in (27)–(28), where $h_v$ and $h'_v$ denote the node embeddings before and after aggregation, respectively. The initial embedding $h_v^{(0)}$ is obtained from the node features $\boldsymbol{x}_{i,j}^{(n)}$, where $\|$ denotes vector concatenation.

$$m_v = \frac{1}{|\mathcal{N}(v)|} \sum_{u \in N(v)} ELU(W_n h_u) \quad (27)$$

$$h'_v = ELU(W_o[\, ELU(W_s h_v) \,\|\, m_v]) \quad (28)$$

Repeated GCN layers progressively incorporate information from multi-hop future rendezvous candidates, yielding the final representation $h_v^{(L)}$ after $L$ layers. The GCN uniformly aggregates successor-node information, with rendezvous-time and due-date features enabling the representation to encode future rendezvous urgency.

### 3) Action

At each decision step, the action selects one of the unvisited patrolling USVs for rendezvous, with its earliest reachable rendezvous node serving as the corresponding candidate. The policy distribution is computed using an actor query that incorporates the current state of the command USV and information about each candidate rendezvous, as shown in Fig. 2(d). The query is combined with the graph representation obtained from the preceding graph network to compute an action score for each candidate. The action scores are computed as defined in (29)–(30) and converted into a policy distribution over all feasible candidates using a softmax function with temperature $T = 1.5$. Here, $x_c$ and $y_c$ denote the current position of the command USV, $\theta_c$ denotes its current heading, and $\theta_u^{arr}$ and $\Delta t_u$ denote the arrival heading and remaining time until rendezvous for candidate $u$, respectively.

$$q_u = [x_c, y_c, \Delta t_u, \sin\theta_c, \cos\theta_c, \sin\theta_u^{arr}, \cos\theta_u^{arr}] \quad (29)$$

$$a_u = \frac{1}{T} MLP_{value}^{actor}(MLP_q^{actor}(q_u) \,||\, h_u^{(L)}) \quad (30)$$

### 4) Reward

The reward at each decision step is defined as the negative sum of the elapsed time until rendezvous completion and the selected USV's tardiness at the rendezvous time. Because the next decision step begins immediately after each rendezvous, the elapsed-time terms accumulate to the total mission completion time, while the tardiness terms accumulate to the total tardiness. Thus, maximizing the cumulative reward is equivalent to minimizing the objective function.

### 5) Critic Network and Policy Optimization

Proximal Policy Optimization (PPO) [13] is employed to stabilize policy updates. The critic uses the same inputs as the actor but estimates the state value with separate parameters. For each candidate node $u \in V_{\text{alive}}$, it computes a gate $e_u$ and value vector $v_u$, whose weighted sum forms a global context vector for estimating $V(s)$, as defined in (31)–(33).

$$e_u = \sigma(MLP_{gate}^{critic}(MLP_q^{critic}(q_u) \,||\, h_u^{(L)})) \quad (31)$$

$$v_u = MLP_{value}^{critic}(MLP_q^{critic}(q_u) \,||\, h_u^{(L)}) \quad (32)$$

$$V(s) = MLP_{readout}^{critic}(\textstyle\sum_{u \in V_{alive}} e_u v_u) \quad (33)$$

## IV. EXPERIMENTS AND RESULTS

All experiments used the same ocean-current generation procedure and USV parameters. A time-invariant current field was generated over the 3600 m × 3600 m operating area by superimposing Lamb–Oseen vortices at large, medium, and small scales to produce smooth spatial variations. The resulting ocean-current velocity at position $\boldsymbol{p}$ is defined in (34). Here, $g \in \{L, M, S\}$ denotes the vortex scale, and $N_g$ is the number of vortices of scale $g$. The vortex induced tangential velocity is given by (35).

$$u_c(\boldsymbol{p}) = \sum_{g \in \{L,M,S\}} \sum_{i=1}^{N_g} u_g^{(i)}(\boldsymbol{p}) \quad (34)$$

$$v_{\theta,g}(r) = \frac{\Gamma_g}{2\pi r}\left[1 - exp\left(-\frac{r^2}{a_g^2}\right)\right],\ \Gamma_g = 2\pi R_g V_g \quad (35)$$

Here, $a_g$ is the core radius, while $R_g$ and $V_g$ are the reference radius and velocity used to determine the circulation strength $\Gamma_g$, respectively. The parameter $d_{\min}$ specifies the minimum separation between vortices. The parameters used for each vortex scale are summarized in TABLE I.

TABLE I. Parameters for multiscale ocean-current generation

| Vortex | $N_g$ | $V_g$(m/s) | $R_g$(m) | $a_g$(m) | $d_{min}$(m) |
|---|---|---|---|---|---|
| $u_L$ | 2-3 | 0.5 | 1800 | 900 | 900 |
| $u_M$ | 3 | 0.4 | 900 | 550 | 450 |
| $u_S$ | 6 | 0.4 | 250 | 350 | 250 |

The vortex parameters were selected to generate current fields with distinct spatial scales and moderate magnitudes relative to the USV cruising speed. The number of large-scale vortices was sampled uniformly, and the vortices were arranged in symmetric pairs or triangular configurations, with the entire pattern randomly rotated. Rotational directions were balanced to avoid directional bias. Across 1,000 generated current fields, the mean current speed was 0.71 m/s, with a variance of 0.15 $(\text{m/s})^2$. The USV cruised at 2.5 m/s with a maximum turn rate of 2.0°/s, giving a turning radius of about 72 m in still water. All experiments were conducted on a workstation equipped with an Intel Core i7 CPU, 64 GB RAM, and an NVIDIA RTX 5070 GPU.

### A. Current-Aware Path Planning with Turn Constraints

#### 1) Grid-Based Path Planning

The maximum turning angle $\theta_{\max}$ was evaluated using a lightweight stochastic tracking procedure to validate the continuous realizability of grid-based paths. For 2,000 Dijkstra paths generated under random current fields, start-goal pairs, and headings, using $R = 60m$ and a 24-direction neighborhood, each path underwent up to 10 stochastic tracking trials to account for controller stochasticity and was regarded as continuously realizable if at least one trial satisfied the 5 m position and 5° heading tolerances. TABLE II reports the feasibility rates for each candidate $\theta_{\max}$. Here, $P_p$ and $P_h$ denote the probabilities of the Path-Proximity and Heading-Alignment strategies, respectively. The setting $[P_p, P_h] = [0.67, 0.33]$ allowed the largest accepted turning angle. Using a lower bound of at least 99% from the 95% Wilson confidence interval as the criterion, $\theta_{\max} = 60°$ was selected, achieving a feasibility rate of 99.45%.

TABLE II. Tracking success rates across policy settings

| $[P_p/P_h]$ | 50° | 55° | 60° | 65° | 70° | $\theta_{max}$ |
|---|---|---|---|---|---|---|
| [0.80/0.20] | 99.65 | **99.45** | 99.35 | 99.30 | 90.05 | 55° |
| [0.75/0.25] | 99.60 | **99.45** | 99.40 | 99.20 | 99.0 | 55° |
| [0.67/0.33] | 99.60 | 99.50 | **99.45** | 99.25 | 99.00 | 60° |
| [0.50/0.50] | 99.30 | 99.15 | 98.80 | 98.60 | 98.55 | - |

For $\theta_{\max} = 60°$, cumulative feasibility rates were 96.4%, 98.85%, and 99.10% within 1, 3, and 5 trials, respectively. Each trial required only 0.38 s on average, making repeated tracking practical. Among successful trials, the minimum successful travel time was used as the continuous-tracking reference. Relative to this reference, the mean absolute percentage error (MAPE) of the grid-based travel times was 1.42%, with a 95th percentile of 5.12%, supporting the validity of the grid-based travel-time approximation.

TABLE III evaluates the effect of grid spacing $R$ under the same experimental conditions. Here, $d^G_{\max}$ and $\rho$ denote the maximum distance to the nearest grid point and grid-point density in $points/km^2$, respectively, while $\bar{\tau}$ denotes the average computation time. As $R$ increased, $d^G_{\max}$ increased and $\rho$ decreased, indicating lower spatial resolution. Meanwhile, the larger feasible $\theta_{\max}$ helped maintain comparable path performance while computation time decreased. Accordingly, $R = 60$m was selected for the multi-rendezvous experiments to balance spatial resolution and computational efficiency.

TABLE III. Grid spacing effects on path quality and spatial resolution

| R(m) | $\theta_{max}$(°) | $T_Q$(s) | $d^G_{\max}$(m) | $\rho$ | $\bar{\tau}$(s) |
|---|---|---|---|---|---|
| 45 | 50 | 829.7 | 26.0 | 570.2 | 8.47 |
| 60 | 60 | 832.3 | 34.6 | 320.8 | 2.49 |
| 90 | 80 | 831.7 | 52.0 | 142.6 | 1.69 |

The neighborhood size and grid geometry were also examined. The 12-direction setting had a smaller feasible $\theta_{max}$ due to shorter transitions and occasionally caused path-search failures, prompting adoption of the 24-direction setting. Triangular and square grids showed comparable performance, with minor trade-offs in spatial resolution and computation time. The triangular grid was used in subsequent experiments.

#### 2) Surrogate Modeling and Path Search

Three regression models were considered: ridge linear regression (LR), extreme gradient boosting (XGB), and random forest (RF). The models were trained on 4,000 random grid paths and evaluated on 1,000 independent paths, with corridor widths set to $3R = 180$m. TABLE IV compares the absolute (Abs) and residual (Res) prediction errors, using Euclidean travel time as a reference and Dijkstra travel time as ground truth. Residual prediction generally reduced error, and XGB-Res achieved the lowest errors for both metrics and was therefore selected.

TABLE IV. Surrogate model comparison

| Metric | LR | | XGB | | RF | | Ref. |
|---|---|---|---|---|---|---|---|
| | Abs | Res | Abs | Res | Abs | Res | |
| MAPE(%) | 10.84 | 8.39 | 3.94 | **3.37** | 4.83 | 3.89 | 18.09 |
| RMSE(s) | 83.30 | 86.52 | 49.84 | **45.46** | 58.10 | 51.87 | 197.86 |

TABLE V compares direct surrogate prediction, A* with Euclidean and surrogate-based heuristics, and Dijkstra in terms of path cost and mean computation time. The mean absolute gap is computed relative to the Dijkstra path cost. Direct prediction is the fastest but does not generate an actual path. Although the heuristics do not guarantee optimality, A* with the Euclidean heuristic achieved only a small path-cost gap while reducing computation time by approximately thirteenfold compared with Dijkstra. Accordingly, A* with the Euclidean heuristic was adopted for subsequent grid-based path searches in rendezvous evaluation.

TABLE V. Comparison of path-planning methods

| Method | Direct Prediction | Euclidean A* | Surrogate-guided A* | Dijkstra |
|---|---|---|---|---|
| Run time (s) | 0.001 | 0.195 | 1.652 | 2.490 |
| Abs. gap (%) | 3.023 | 2.113 | 0.883 | 0.0 |
| Path output | No | Yes | Yes | Yes |

### B. Graph-Based RL for Rendezvous Planning

For the multi-rendezvous problem, two patrol-route types were considered for each fleet size, with the initial position, patrol direction, and priority of each USV randomly assigned. Type A comprises non-overlapping rectangular routes, whereas Type B comprises overlapping routes: four horizontal rectangles, four vertical rectangles, and four squares for 12 USVs, with four additional squares for 16 USVs. For 24-USV instances, both types used 12 routes, with two USVs assigned to each route at opposite starting positions. Each route was formed by sequentially connecting 18 major waypoints placed at fixed offsets from the area boundaries. A common due-date reference interval was obtained by dividing the mean rendezvous completion time of priority-agnostic greedy solutions over randomly sampled instances by the number of priority tiers. This interval was applied to all instances of that type to assign priority-dependent due dates.

Training used 12 patrolling USVs, three priority levels, and a mixture of Types A and B. The learning rate, discount factor γ, GAE parameter λ, clipping ratio ε, and value-loss coefficient were set to $2.5 \times 10^{-4}$, 1.0, 0.95, 0.2, and 0.5, respectively, with no entropy bonus. Each of 500 iterations generated six instances with eight rollouts each, yielding 48 trajectories trained for two epochs. During training, the surrogate model determined transition times and reachability, while the arrival heading was approximated by the bearing to the rendezvous point. During evaluation, the sFTG encoded the state. For each unvisited USV, candidate rendezvous locations were identified and ordered using surrogate estimates and checked sequentially by grid-based path planning until a feasible location was found. Grid-based path planning used the current heading, and the resulting arrival heading was propagated to the next decision. TABLE VI shows the network architecture.

TABLE VI. Graph-RL network architecture

| Component | Architecture | Activation |
|---|---|---|
| Node embedding | Linear (4,128) | - |
| Graph encoder | 3 layers ($d_h = 128$) | ELU |
| $MLP_{q}^{actor/critic}$ | MLP(7, 64, 128) | ELU |
| $MLP_{value}^{actor}$ | MLP(256, 128, 64, 1) | ELU |
| $MLP_{gate}^{critic}$ | MLP(256, 128, 64, 1) | ELU, sigmoid |
| $MLP_{value}^{critic}$ | MLP(256,128,128) | ELU |
| $MLP_{readout}^{critic}$ | MLP(128,64,1) | ELU |

For direct comparison, MLP-RL removes the graph encoder from the proposed Graph-RL and independently scores candidates using query and node features. Greedy-IT considers incremental tardiness of all unfinished USVs during a transition, whereas Greedy-DT considers only the selected USV's tardiness. Priority-Constrained Nearest Neighbor (PCNN) selects USVs by priority and then proximity. As a stronger heuristic, 2-step lookahead greedy (2-step LG) extends Greedy-IT to two-step sequences and executes the first action of the best sequence. Exhaustive evaluation over $N!$ rendezvous orders is costly because each sequence requires repeated path searches. Greedy-IT and MLP-RL therefore serve as the primary baselines, while 2-step LG provides a stronger, tractable search-based comparison. Each problem setting used 60 test instances, comprising 30 each of patrol Types A and B, with stochastic RL results averaged over 10 runs per instance. The objective was normalized by 2,000 s.

TABLE VII reports Graph-RL and MLP-RL results from three seeds under the same training setting. The mean and standard deviation are calculated from the mean objective values obtained for each seed. Both methods showed low variation across seeds, while Graph-RL consistently outperformed MLP-RL under both surrogate- and grid-based evaluations. Given the low seed-to-seed variation, further evaluations used a near-average seed from TABLE VII on a separate test set.

TABLE VIII reports the mean and standard deviation across test instances for each patrol fleet sizes $N$ and priority level $T$. Dark and light shading indicate $p < 0.05$ and $p < 0.1$, respectively, in paired t-tests against MLP-RL. Boldface denotes the lowest mean objective value. Relative improvements were computed by averaging instance-wise relative objective reductions.

TABLE VII. Performance variability across training seeds

| Algorithm | Surrogate | | Grid-graph | |
|---|---|---|---|---|
| | Mean | Std | Mean | Std |
| Graph-RL | 4.145 | 0.017 | 4.258 | 0.066 |
| MLP-RL | 4.333 | 0.004 | 4.400 | 0.048 |

Graph-RL reduced the objective by 3.50% and 10.52% on average compared with MLP-RL and Greedy-IT, respectively, both statistically significant at the 5% level. Despite not explicitly encoding future rendezvous relationships, MLP-RL and Greedy-IT remained competitive baselines. Compared with the 2-step lookahead greedy across all test instances, Graph-RL significantly reduced the objective by 1.48% at the 5% level while requiring only about 1/35 of the computation time. The higher runtime of 2-step LG results from evaluating $O(N^3)$ two-step candidate sequences, making additional lookahead steps computationally demanding. Graph-RL and MLP-RL showed similar computation times, while Greedy-IT had an intermediate runtime and Greedy-DT and PCNN were faster. Path search dominated computation time for all methods. Runtime differences among the methods became more pronounced under the 3T and 4T settings, with the gap between the RL methods and the faster heuristics increasing to approximately 2.5 times that observed at 2T. Although priority does not directly affect computational cost, the resulting rendezvous sequences may lead to longer distances to subsequent candidates, increasing path-search costs.

TABLE *IX* shows that the advantage over MLP-RL grew with fleet size under both evaluation methods. As larger fleets introduce more potential rendezvous relationships, this trend suggests an increasing benefit from encoding future connectivity rather than encoding candidates independently. Across priority tiers, both evaluations showed the same ordering, with the largest improvement at 2T and smaller gains at 3T and 4T, likely reflecting the relatively stronger performance of MLP-RL as the improvement over Greedy-IT remained stable. The similar magnitudes across both evaluations suggest that the sFTG remains informative despite approximation and that the benefit of graph encoding is preserved.

Fig. 3 shows representative multi-rendezvous trajectories. Blue, orange, and green indicate patrol routes in descending order of priority, while the green pentagon and red × mark the command USV's start and end points. The policy tends to visit high-priority USVs earlier while also favoring nearby rendezvous with little extra travel time.

TABLE VIII. Objective values and runtime across fleet sizes and priority settings

| Algorithm | N = 12 | | | N = 16 | | | N = 24 | | | Mean runtime |
|---|---|---|---|---|---|---|---|---|---|---|
| | 2T | 3T | 4T | 2T | 3T | 4T | 2T | 3T | 4T | |
| Graph-RL | **3.84 ± 0.7** | **4.28 ± 0.9** | **4.65 ± 1.1** | **4.56 ± 0.7** | **5.63 ± 1.1** | **6.32 ± 1.4** | **7.40 ± 2.4** | **7.95 ± 1.4** | **9.25 ± 1.7** | 105.37 s |
| MLP-RL | 3.86 ± 0.6 | 4.41 ± 0.8 | 4.72 ± 0.8 | 4.87 ± 0.8 | 5.80 ± 1.0 | 6.55 ± 1.2 | 7.85 ± 2.2 | 8.40 ± 1.2 | 9.76 ± 1.5 | 106.63 s |
| Greedy-IT | 4.42 ± 1.2 | 5.17 ± 1.5 | 5.63 ± 1.7 | 5.11 ± 1.1 | 6.37 ± 1.4 | 6.98 ± 1.5 | 9.39 ± 3.8 | 9.63 ± 2.5 | 11.24 ± 2.9 | 90.22 s |
| Greedy-DT | 5.27 ± 1.5 | 6.81 ± 1.6 | 7.96 ± 2.0 | 6.25 ± 1.6 | 9.34 ± 2.3 | 10.47 ± 2.5 | 14.21 ± 6.4 | 15.49 ± 3.6 | 18.61 ± 4.2 | 70.30 s |
| PCNN | 4.05 ± 0.7 | 5.05 ± 1.3 | 6.42 ± 2.1 | 4.79 ± 0.9 | 6.93 ± 1.8 | 9.84 ± 4.0 | 9.67 ± 5.1 | 10.38 ± 3.1 | 16.43 ± 6.1 | 66.68 s |
| 2-step LG | 4.18 ± 1.0 | 4.53 ± 1.0 | 4.89 ± 1.1 | 4.94 ± 1.2 | 6.11 ± 1.8 | 6.65 ± 2.0 | 8.07 ± 2.4 | 8.34 ± 2.1 | 9.69 ± 2.4 | 3737.2 s |

TABLE IX. Relative improvements over primary baselines under grid-based and surrogate-based evaluation

| Comparison | Evaluation | Fleet size | | | Priority tier | | | Macro avg. (%) |
|---|---|---|---|---|---|---|---|---|
| | | N = 12 | N = 16 | N = 24 | 2T | 3T | 4T | |
| Graph-RL vs. MLP-RL | Grid graph | 1.29 | 3.86 | 5.35 | 3.79 | 3.44 | 3.27 | 3.50 |
| | Surrogate | 2.91 | 3.82 | 5.08 | 4.88 | 4.50 | 2.44 | 3.94 |
| Graph-RL vs. Greedy-IT | Grid graph | 10.03 | 7.46 | 14.06 | 10.09 | 11.24 | 10.22 | 10.52 |
| | Surrogate | 8.59 | 11.00 | 17.46 | 11.62 | 12.54 | 12.88 | 12.35 |

(a) 12 USVs, 3T, Type A (b) 12 USVs, 3T, Type B (c) 16 USVs, 3T, Type A

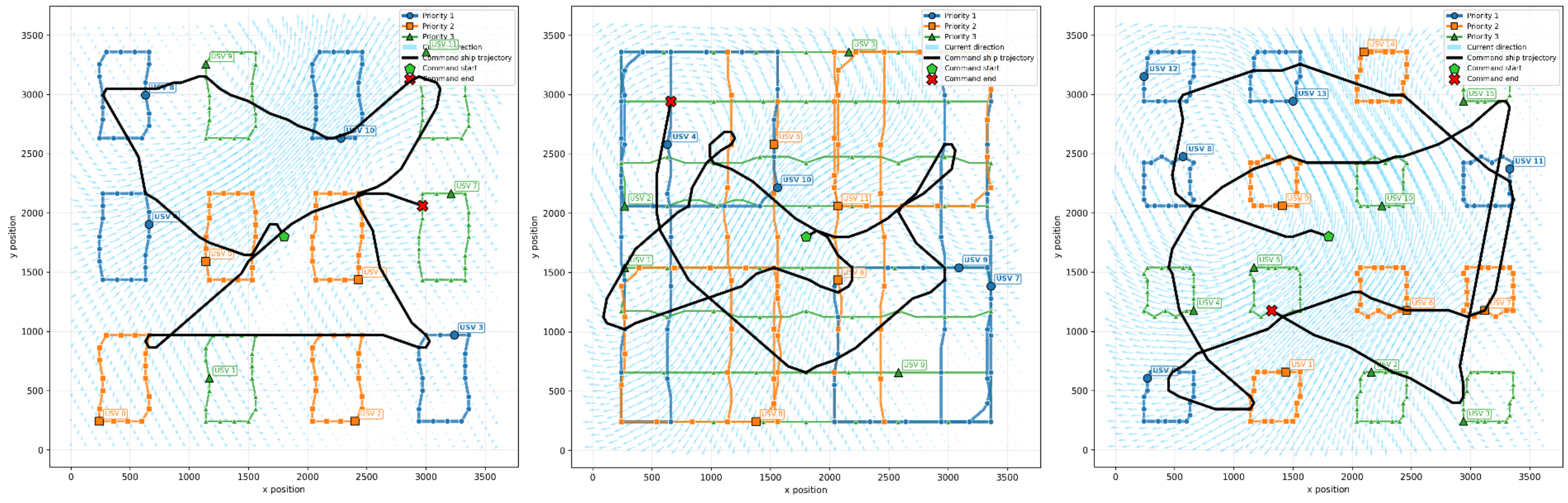

Fig. 3. Representative rendezvous trajectories generated by Graph-RL

## V. CONCLUSION

This study proposed a computationally tractable framework for multi-rendezvous planning under ocean-current and turn-rate constraints. Continuous turn-rate constraints were approximated by maximum turning-angle constraints calibrated through trajectory-tracking simulations, while a surrogate travel-time model enabled efficient construction of the sFTG for GCN-based rendezvous policy learning.

The proposed policy outperformed Greedy and MLP-RL across various problem settings and showed a statistically significant improvement over lookahead greedy while requiring only about 1/35 of its computation time. The study demonstrates that surrogate-based state representation and policy learning remain effective under evaluation with explicit path and arrival-heading computation. Nevertheless, trajectory-tracking validation over complete multi-rendezvous missions and the effects of assuming a fixed command USV heading during waiting remain to be examined. Overall, the proposed framework provides an effective approach to complex multi-rendezvous planning under ocean currents and turn-rate constraints and can serve as a basis for future studies incorporating more realistic mission-level simulations.